# Combining Machine Learning with Knowledge Engineering to detect Fake News in Social Networks-a survey

Sajjad Ahmed[1], Knut Hinkelmann[2], Flavio Corradini[1]

[1]Department of Computer Science, University of Camerino, Italy
[2]FHNW University of Applied Sciences and Arts Northwestern Switzerland
Riggenbachstrasse 16, 4600 Olten, Switzerland
ahmed.sajjad@unicam.it; knut.hinkelmann@fhnw.ch; flavio.corradini@unicam.it

**Abstract**

Due to extensive spread of fake news on social and news media it became an emerging research topic now a days that gained attention. In the news media and social media the information is spread high-speed but without accuracy and hence detection mechanism should be able to predict news fast enough to tackle the dissemination of fake news. It has the potential for negative impacts on individuals and society. Therefore, detecting fake news on social media is important and also a technically challenging problem these days. We knew that Machine learning is helpful for building Artificial intelligence systems based on tacit knowledge because it can help us to solve complex problems due to real word data. On the other side we knew that Knowledge engineering is helpful for representing expert's knowledge which people aware of that knowledge. Due to this we proposed that integration of Machine learning and knowledge engineering can be helpful in detection of fake news. In this paper we present what is fake news, importance of fake news, overall impact of fake news on different areas, different ways to detect fake news on social media, existing detections algorithms that can help us to overcome the issue, similar application areas and at the end we proposed combination of data driven and engineered knowledge to combat fake news. We studied and compared three different modules text classifiers, stance detection applications & fact checking existing techniques that can help to detect fake news. Furthermore, we investigated the impact of fake news on society. Experimental evaluation of publically available datasets and our proposed fake news detection combination can serve better in detection of fake news.

## Introduction

Fake news and the spread of misinformation have dominated the news cycle after US Presidential elections in 2016. Some reports show that Russia has created millions of fake accounts and social bots to spread false stories during the elections (Lewandowsky 2017).Various motivations are observed for spreading fake news and generating these types of information on social media channels. Some of them are to gain political gains or ruin someone else's reputation or for seeking attention. Fakenews is a type of yellow journalism or propaganda that consists of deliberate misinformation or hoaxes spread via traditional print and broadcast news media or online social media.

The importance of fake news can easily be understood as per the report published by PEW Research Centre (Rainie et al. 2016). The statistics shows that 38% of adults often get news online, 28% rely on website/apps & 18% rely on social media. Overall 64% of adults feel that fake news causes a great deal of confusions. The importance of fake news can also be judged through below diagram shows dramatically fake news gained worldwide popularity in 2016 after US presidential elections.

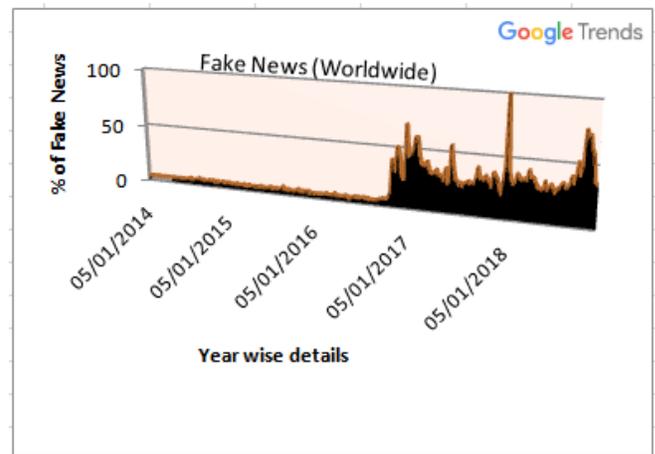

*Figure 1: Last five years Google trend*

The wide-ranging spread of fake news can have a negative impact on society and individuals. Fake news intentionally persuades clients to accept biased or false beliefs. Fake news changes the way people interpret and respond to real news. For example, some fake news was just created to trigger people's distrust and make them confused, impeding their abilities to differentiate what is true and what is not true (Scott et al., 2000; Leonhard et al., 2017; Himma 2017). It is important to understand that fake and deceptive news have existed for a long time. It's been part of the conversation as far as the birth of the free press (Soll 2016). There are various approaches for automated fake news detection: Text Classification, Stance Detection, Metadata & Fact Checking.

## Data Driven:

- Text classification: They mainly focus on extracting various features of text and after that incorporating of those features into classification models e.g. Decision tree, SVM, logistic regression, K nearest neighbor. At the end selection of best algorithm that performs well (Nidhi et al. 2011). Emergent[1] is a real time data driven rumor identification approach. It works automatically to track rumors associated with social media but those rumors where human input require has not been automated. The problem is that most classification approaches are supervised so we need prior dataset to train our model but as we discussed that obtaining a reliable fake news dataset is very time consuming process.
- Stance detection: False news has become an important task after the US 2016 Presidential elections. Governments, Newspapers and social media organizations are working hard to separate the fake contents and credible. So the first step in identification phase is to understand that what others are saying about the same topic (Ferreira et al. 2016). So far as fake news challenge initially to focus on stance detection. In stance detection we check the estimation of relativity of two different text pieces on the same topic and stance of others (Saif et al. 2017). PHEME[2] was a three years research project funded by the European Commission from 2014-2017, studying natural language processing techniques for dealing rumor detection, stance classification (Lukasik et al. 2015 ; Zubiaga et al. 2016), contradiction detection and analysis of social media rumors. Existing stance detection approaches based on embedding features on individual posts to predict stance of that particular content.
- Meta-data: We can analyze fake news differently with different measure similarities e.g. Location, Time, author and Quality. we can detect whether the same news published by other media agencies or not, We can check the location of the news Maybe a news has a higher probability of being fake, if it is generated somewhere else and not at the location they deal with (e.g. Trump writes about China or Arabian States, News about Hillary Clinton has its origin in Russia), We can check news quality wise it is more probable that fake news do not have mentioned their sources, simply claim something, while for real news the source is mention and also we can check the time of the news as whether the same news appears in other media or sourced if it is repeated more often in the beginning, because they are interesting, and become recognized as fake with the time, which reduces the repetition or they are deleted from some websites. At this stage we don't have definitive solution but after detailed literature review we can say that it's true that producing more reports with more facts can be useful for helping us to make such decisions and find technical solutions in fake news detection.

## Knowledge Engineering:

- Fact Checking techniques mainly focus on to check the fact of the news on the basis of the known facts. There are three types of fact checking techniques available Knowledge Linker (Ciampaglia et al. 2015), PRA (Lao et al. 2011), and PredPath (Shi et al. 2016). Then the Predictions algorithms that using knowledge to check the fact are Degree Product (Shi et al. 2016), L. Katz (1953), Adamic & Adar (Adamic et al. 2003) and Jaccard coefficient (Liben et al. 2016). Some fact checking organizations providing online fact checking services e.g. Snopes[3], PolitiFact[4] & Fiskkit[5]. Hoaxy[6] is another plate form for fact checking. Collection, detection and analysis and to check online misinformation is part of Hoaxy. The criteria they followed is to check the news is fake or not fake simply they refer it to the domain experts, individuals or organizations on that particular topic. They also followed non partisan information and data sources (e.g., peer-reviewed journals, government agencies or statistics).

## Discussion

Our main research question that how would one distinguishes between fake and non fake news articles using data driven and knowledge engineering. The facts show that the fake news phenomenon is an important issue that requires scholarly attention to determine how fake news diffused. Different groups introduced different models some of them applied data oriented and some applied only knowledge side. The important point is the speed of spreading of these types of information on social media networks is challenging problem that require attention and alternative solution. If news is detected fake the existing techniques blocked them immediately due to their functionally as we can't replace them but if a news detected fake at least we need some experts opinion or verification before blocking that particular news. This thing helps to rise the third party fact checking organizations to come and solve the issue but that is also time consuming process. We need some application that check the news whether it is fake or not at the same place.

---

[1] www.emergent.info
[2] www.pheme.eu

[3] www.snopes.com

[4] www.politifact.com

[5] www.fiskkit.com

[6] https://hoaxy.iuni.iu.edu/

The existing fake news systems based on the predictive models that simply classify that the news is fake or not fake. Some models used source reliability and network structure so the big challenge in those cases is to train the model due to non availability of corpora it is impossible.

Due to surge of fake news problem and overcoming the challenges discussed above a volunteer based organization FakeNewChallenge[7] that contains 70 teams which organizes specifically machine learning competitions to the detection of fake news problem.

At the end we can say that there is a need of an alternative application that combine knowledge with data and automation of fact checking is required which looks content of the news deeply with expert opinion at the same place to detect the fake news.

The rest of this paper is divided into four sections. Section 2 contains background, impact on society, News content models, related work and similar application areas; Section 3 describes Methodology, Proposed combination approach and publically available data set we used for initial classification. Our conclusions and future directions are presented in Section 4.

# Literature Review

In this section we try to cover all the topics that are related to our topic and can be helpful for better understanding of fake news detection. At the beginning we discuss that the trust level of the readers on online news media. Then we discuss the impact of fake news on society and then different types of news models. Then we discuss related work and similar applications areas where some of the researchers applied data driven and some applied knowledge side to overcome the specific problem in that particular domain.

## Background

Internet gave opportunity to everyone enter online news business because many of them were already rejected the traditional news sources that had gained high level of public trust and also credibility of the work. According to a survey general trust on mass media collapsed as lowest in the history of this business. Especially in political right 51% democrats and 14% republican in USA expressing a great deal and trust in mass media as news source (Lazer et al. 2018).

It has come to known that the information repeated again is more likely to be rated true than the information that has not been heard before. Familiarity with false news would increase with truthfulness. Further this thing did not stop here as the false stories would result to create the false memory. The authors first observed "illusory-truth effect" and gave the results that subject rated repeated statements truer as compare to the new statements. They present a case study with the results that the participants who had read the false news or stories consecutively five weeks believe false stories more truthful and more plausible as compare to the participants who had not been exposed (Hasher et al. 1977).

News can be true if the information it expresses that is more familiar. Familiarity means automatic consequences of exposure so it will influence on truth and that is fully unintentional. In those cases where the source or the agency that circulated stories warns that source may not be credible, people did not stop to believe on that story due to the familiarity (Begg et al. 1992). Another study that contains half statements showing in the experiments were true and half were false but the results shows that the participants like repeated statements although they were false but due to the familiarity they rated as more true than the stories they heard first time (Bacon et al. 1979). Monitoring of source is itself an ability to check and identify the news origin we read. Some studies clearly indicate that the participants use familiarity to understand the source of their memory. Another study that proposed general knowledge and semantic memory does not focus on conditions but it only helps a person when and where he learned this information. Similarly a person may have some knowledge about an event but not remember the event so it comes from memory (Potts et al. 1989).

## Overall Impact on different areas

News is a real time situation and a comprehensive story that covers different issues like criminology, health, sports, politics, business etc. Local news agencies mostly focus on the specific regional issues and international news agencies covers both local and global news. Finding a particular story on the basis of reader's choice is an important task. Different methods proposed in this study that how can we overcome the issue and follow the reader's choice (Zhai et al. 2005). Hot topic detection in a local area during a particular period based on micro blogs containing difference in words but pointing towards same topic using twitter and Wikipedia.

## Business

In online news media the services and total number of users are important to gain more business. Some big names who are earning a lot due to high number of users and circulation of fake news are Facebook, Twitter, Google and Search engines also fake news producers and consumers. Fake news growing dramatically day by day and its impact on society is very bad.

## Social Networks

After US Presidential elections social media facing pressure from general public and civil society to decrees fake news on their platforms. It's a very difficult task to combat fake news and especially when no proper check and balance and sharing policies are available. Articles that go viral on social media can draw significant revenue through advertising when users click and directly redirected on that page. But the question is how we can measure the importance of social media networks for fake news suppliers so one possibility to measure through the source of their web traffic. Every time when a user visits the web page that user has navigated directly through server or it referred to some other site (Allcott et al. 2017).

---

[7] http://www.fakenewschallenge.org/

One focused area that really helpful to detect fake articles on Facebook is the fact checking organizations. According to the Facebook that they are taking all steps to overcome the issue on their platform and make it as much as difficult as possible to buy ads on platform for people who really want to share fake contents. Better identifying false news with the help of community and third party fact checking organizations and some stance detections mechanisms is possible because they can limit the spread speed of fake contents and they can make it uneconomical (Mosseri 2017).

Single users have the same facility that they will get a message that some people do not agree with the article content. The regular users are not in a position to judge the validity of the links they can see. So this thing might be unreliable for Facebook flagging functions (Wohlsen 2015).

The second focused area is that a flag that is available with the fake news article. Simply users can click upper right corner of that post. The more times that particular post flagged by the users that it is false then less often it will show up in news feed tab. According to the Facebook policy that they would not delete that flagged post but they end up with disclaimer with the statement "Many people on Facebook have reported that this story contains false'' Stanford History Education Group (2016).

Due to the sensitivity of the issue Facebook sends a flagged post to the third party who is responsible to check the fact about that post. If fact checking organizations marked it disputed then automatically users will see a banner under the article if it appears in user news feed area. That banner will clearly explain the situation that third party organization disputed it and a link is available. Another thing is that those disputed stories pushed down in news feed and a message will appear before sharing from any user that if they sure about it then they can share it (Guynn 2017). Relying on users is not a permanent and good solution, but the idea is just to educate the users and if they consent on it then they can share it. If every user take care about this then fake news would not be as big problem.

To check the level of truth in articles is very difficult as they differ in some points but in a very professional manner. That's why only the best way is to understand that the management of Facebook that they need to educate their users about sharing policy. Every user need to understand that before sharing any information on Facebook they must be sure about it (Dillet 2017). Facebook management and responsible persons claims that they have an algorithm that helping by rooting out fake articles. The algorithm shows the users about that article before sharing, source, date, topic and number of interactions.

When we compare with Twitter, fake news shared by the real account holders to some small websites and highly active 'cyborg' users (Silva et al. 2016). They are very professional and sometimes these professional groups evolved to be industrialized by states and terrorist organizations. These groups called Troll farms and according to one study that they have potential algorithm to track down in Twitter (Nygren et al. 2016).

## Security Agencies

Misinformation or propaganda has always been used to affect people and create fear for opponent. We can categorize it in three types. White propaganda is that where we knew the initiator and the news circulated by that particular person or group is true. Black propaganda is that where we don't know the source and also the news shared by that person or groups is totally false.

Grey type is that which is between the white and black. During the cold world war the objective of these types of activities is to sway the opinions just to hide and distorted facts from hidden senders. A big example of this type of propaganda happened from 2002 to 2008 when United States Military Department recruited approximate seventy five pensioned officers just to propagate on media on Iraq's possible ownership of weapons. The objective of this activity is to weaken the public of the opponent who supports them and strengthen the own support. The work had done through different sources e.g. radio, newspapers and TV channels that hide the connections (Nygren et al. 2016).

When we compare this with the earlier variants of propaganda due to society needs because today it is possible for everyone to reach large audience within seconds but in past it was not possible. So it means we are more reliant on information that affects more. Some other actors also involved in this campaign and they can easily affect the facts, those are diplomatic persons, military economic state actors and public relation departments. An independent body can easily control these types of activities as compare to the state controls everything. A big example of this disinformation is Ukraine's crisis 2014, where a state invaded another country territory and misleads about it. Due to this it affects badly the world's response. We knew that this is not only one thing that spreading lies but also other activities that linked to it are involved. In next section we discussed different types of news content models one by one with examples.

## News Content Models

In Content modeling we identify our requirements, develop taxonomy (Classification system) that meets those requirements and consider where metadata should be allowed or required.

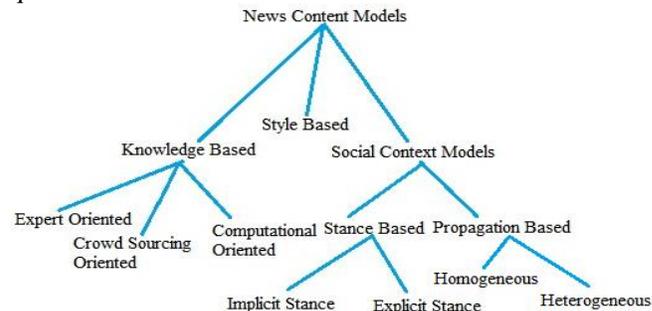

*Figure 2: News Content Models*

News content models can be categories in knowledge based and style based but due to enhancement in social media it provides additional resources to the researchers to supplement and enhance news content models like Social Context Models, Stance Based & Propagation Based. The main focus of news content modeling is on news content features and especially factual sources to detection of fake and real text (Wang 2017). In next section we discussed news content models and existing applications comes under their domain with examples.

### Knowledge-based:

The objective of Knowledge based approach is to use external sources to fact check in news content and the goal of fact checking is assign a truth value to a claim in particular (Vlachos et al. 2014). When we read literature it has come to our knowledge that fact checking in fake news detection area gained high attention. That's the reason many efforts have been made to develop some feasible automated fact checking systems. Since fake news attempts to spread false news contents on social media networks and also news media, so straightforward means detection of those false claims and check the truthfulness of those news. We can categorize existing fact checking applications in three parts expert oriented, crowd sourcing oriented and computational oriented.

- **Expert Oriented**

We need highly domain experts in expert oriented fact checking that can investigate data and documents to verdict the claims. The famous fact checking applications are Snopes[8] & PolitiFact[9]. Expert oriented fact checking is very demanding but it's also time consuming process. As soon as they receive new claim they consult domain experts, journals or statistical analysis already available in that particular domain. It took so much time so we need to develop a new classification approach that can help to detect fake news in a better way and timely.

New fact check mechanisms that can help readers after critically evaluate the news before judgment by using fact checking. The objective of this work is not to provide results that the content is fake or not fake instead of provide mechanism for critically evaluation during news reading process. Reader starts reading of the news and fact check technique will provide facility to the reader that at the same time read all related or linked stories just for critical evaluation. They used scoring measure formula that displays the related stories of the scoring measure threshold but if the scoring measure below the threshold it will not display on corresponding fact check page (Guha 2017).

Three generally agreed upon characteristics of fake news: Text of an article, user response and the source that needs to be incorporate at one place and after that they proposed a hybrid model. First module captures the abstract temporal behavior of the users, measure response and text. Then the second component score estimates source for every user and then combined with the first module (Ruchansky et al. 2017). At the end the proposed model allows CSI to output prediction separately shown in figure-3.

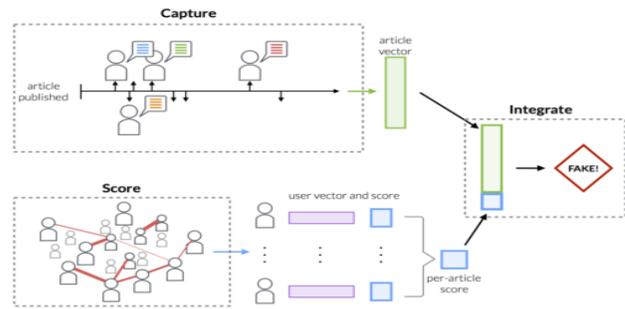

*Figure 3: Illustration of proposed CSI model*

- **Crowd sourcing Oriented**

In crowd sourcing approach it gives option to the users to discuss and annotate the accuracy of specific news. So in other words we can say it's fully rely on the wisdom of crowd to enable fact checking on the basis of their knowledge. Fiskkit[10] is a big example of this type of fact checking as it provides facility to the users to discuss and annotate the accuracy of news article. Another anti fake news detection application that provide facility to detect fake articles and further it gives facility to users to report suspicious news contents so that the editors will check it further. After taking motivations from Facebook flag method with the involvement of public and leveraging crowd signals for detecting fake contents (Potthast et al. 2016). An algorithm named Detective was developed as it checks run time flagging accuracy with Bayesian inference method. This algorithm selects small subsets of news everyday and send back to the expert and on the basis of expert response it stops that fake news.

- **Computational Oriented**

Computational fact checking aims to provide users an automatic system that can classify true and false contents. Mostly computational fact checking works on two points that identify check worthy claims and then discriminate the veracity of fact claims.

It works on the key basis and viewpoints of users on the specific content (Houvardas et al. 2006). Open web and structured knowledge graphs are the big examples of these types of computational oriented fact checking. Open web sources are utilized as referenced that can differentiate the news true and false (Banko et al., 2007; Magdy et al., 2010).

Separation of fake contents in three categories: serious fabrication, large scale hoaxes and humorous fake was the main objective of this work. They provide a way to filter, vet and veri-fying the news and discussed in details the pros and cons of those news (Rubin, V et al., 2015).

---

[8] www.snopes.com

[9] www.politifact.com

[10] www.fiskkit.com

This study is data oriented application simply they used available dataset and then applied deep learning method and finally they proposed a new text classifier that can predict whether the news is fake or not (Bajaj 2017). Dataset used for this project was drawn from two different publically accessible websites[11][12]

Traditionally all rumor detection techniques based on message level detection and analyzed the credibility on the basis of data but in real time detection based on the keywords then the system will gather related micro blogs with the help of data acquisition system which solves this problem.

The proposed model combines user based, propagation based and content based models and check real time credibility and sends back the response within thirty five seconds (Zhou et al. 2015).

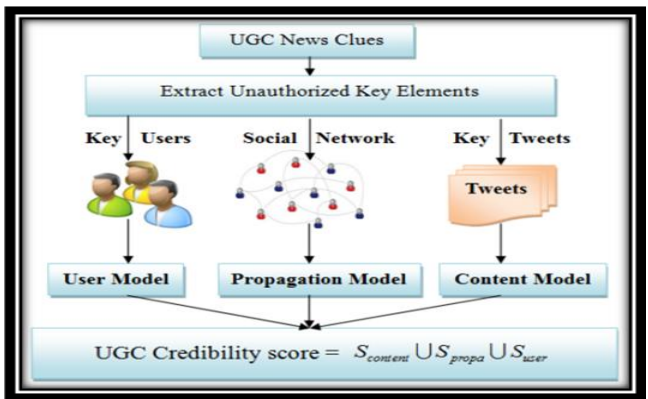

*Figure 4: Framework of Real time Rumor Detection*

## Style-based:

In style based approach fake news publishers used some specific writing style necessary to appeal a wide scope that is not available in true news article. The purpose of this activity is to mislead or distorted or influence large population.

Categorization of News sources into two categories: writing quality and strong sentiment is the main point as real news sources have higher writing quality (taking into account: Misspelled Words, Punctuation & sentences length) as compare to the fake news articles that are likely to be written by unprofessional writers. On the other side real news sources are appear unbiased or neutral words, describing events with facts. So the development of classifier and compare it with other classification methods is the main focus area for fake content identification (Fan et al. C 2017).

It is hard to pin down satire in the scholarly literature (Nidhi et al. 2011). Another study that proposed a method that can first translated the theories of humor, irony and satire into a predictive method for satire detection. Conceptual contributions of this work are to link satire, irony and humor. Then target the fake news frames with filtering due to its potential of mislead to the readers (Rubin et al. 2016).

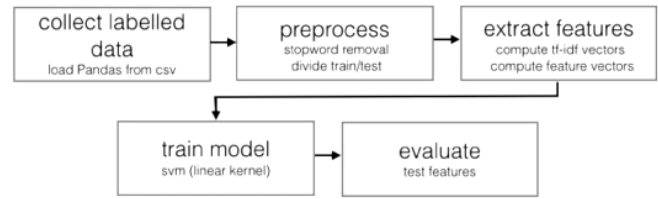

*Figure 5: Fake News: Satire Detection Process*

## Social Context Models:

Social media provides additional resources to the researchers to supplement and enhance news context models. Social models engagements in the analysis process and capturing the information in different forms from a variety of perspectives. When we check the existing approaches we can categories social modeling context in stance based and propagation based. One important point that we need to highlight here that only a few existing social context models approaches utilized for fake news detection. So we will try with the help of literature those social context models that used for rumor detection. Proper assessment of fake news stories shared on social media platforms and identification of fake contents automatically with the help of information sources and social judgment on the basis of Facebook data is the main point of this work. During 2016 US President Elections they examines that machine learning classifiers can be helpful to detect fake news (Tresh et al. 1995).

- **Stance-based**

It is a process that can determine the results from news that the reader is in favor or against or neutral of that particular news (Saif et al. 2017). There are two ways to represent the user stances explicitly or implicitly. Explicit stances are those stances where the readers gave direct expressions like thumb up or thumb down. Implicit stances are those stances where results extracted from social media posts. Overall we can say that stance detection is a process where automatically determining from user posts that's the majority of users or in favor or against (Qazvinian et al., 2011; Jin et al., 2016) proposed a model to check the viewpoint of users and then on the basis of viewpoint to learn the credibility of posts. (Tecchini et al. 2016) proposed a bipartite network of users on Facebook posts using 'like' stance. On the basis of the results we can predict likelihood of Facebook users.

Stance detection of headlines based on n-gram matching for binary classification "related" vs. "unrelated" pairs. This approach can be applied detection of fake news especially clickbait detection. They used dataset released by the organization Fake News Challenge (FNC1) on stance detection for experiments (Bourgonje et al. 2017). The dataset is publically available and can be downloaded from the corresponding GitHub page

---

[11] www.kaggle.com

[12] https://research.signalmedia.co/newsir16/signal-dataset.html

along with base line implementation. Key points of the dataset can be seen in the below figure-4.

| Unique headlines | 1.648 | |
|---|---|---|
| Unique articles | 1.668 | |
| Annotated pairs | 49.972 | 100% |
| Class: unrelated | 36.545 | 73% |
| Class: discuss | 8.909 | 18% |
| Class: agree | 3.678 | 7% |
| Class: disagree | 840 | 2% |

*Figure 6: Key Points of the FNC1 dataset*

- **Propagation-based**

In propagation based approach homogeneous and heterogeneous credibility networks built for propagation. Homogeneous propagation that contains single entities like post or event but heterogeneous credibility network contains multiple entities like posts, events and sub events (Jin et al 2016; Zhiwei et al 2014; Gupta et al 2012). In propagation based approach we check the interrelation of relevant events on social media posts to detect the fake news and the credibility of that news. Another study that helps to build three layer network after including sub events then we can check the credibility of news with the help of graph optimization framework (Jin et al. 2014). Propagation based algorithm for users encoding that can check the credibility and tweets together (Gupta et al. 2014)

## Similar Application Areas

In this section we will discuss similar application areas to the problem of fake news detection. Some applications used data side and some are related to the knowledge side. They perform good results in specific domain but they require high efforts during development so the combination with knowledge engineering it can be helpful to reduce the efforts. At the end we discussed some other data driven applications (table-1) and few where the combination of data driven and knowledge exists (table-2).

## Truth Discovery/Hot Topic Detection

Truth discovery plays a distinguished role in information age as we need accurate information now more than ever. In different application areas truth discovery can be beneficial especially where we need to take critical decisions based on the reliable information extracted from different sources e.g. Healthcare (Yaliang et al. 2016), crowd sourcing (Tschiatschek et al. 2018) and information extraction (Highet 1972). Some cases we have the information but we are unable to explain so those cases knowledge engineering can take part and we can better predict as per the learning from the previous results.

## Rumor Detection

Objective of Rumor detection is to classify a piece of information as rumor or non rumor. Four steps are involved model Detection, Tracking, Stance & Veracity that can help to detect the rumors. These posts considered the important sensors for determining the authenticity of rumor. Rumor detection can further categories in four subtasks stance classification, veracity classification, rumor tracking, rumor classification (Arkaitz et al. 2017). So still few points that require more details to understand the problem and also we can learn from the results that is it actually rumor or not and if its rumor then how much it is. So for these questions we believe that combination of data and knowledge side is required to explore those areas that still unexplainable.

## Clickbait Detection

Attract visitor's attention and encourage them clicking on a particular link is the main objective in clickbait business. Existing Clickbait approaches utilize various extraction features from teaser messages, linked WebPages, tweets Meta information (Martin et al. 2016). So in same case we can notify the readers before reading any kind of news that it could be fake due to some specific indications so the readers need to be more careful.

## Email Spam Detection

Spam detection in email is one of the major problem that bringing financial damage to the companies and also annoying individual users. Different groups are working with different approaches to detect spam in email and different machine learning approaches are very helpful for spam filtering.

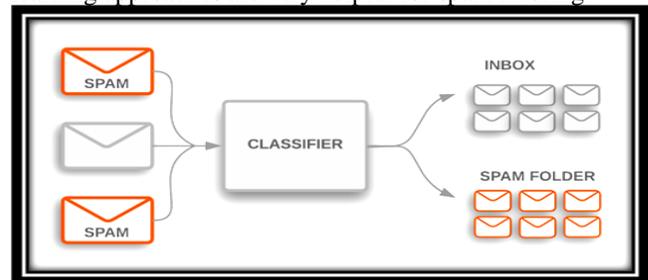

*Figure 7: Spam Filtering*

Spam causes different problems as broadly we discussed on top but more precisely spam causes misuse of traffic, computational power and storage space. This study also explains that many other different techniques can be helpful for spam detection like Email filtering, Blacklists unauthorized addresses, White lists, Legal actions and many more (Siponen et al. 2006).
Below in two tables we just gave an overview of data and knowledge side and specific application domain where they applied to resolve the issue.

| Data Driven systems/applications | Authors | Year |
|---|---|---|
| Data Driven Modelling framework For Water Distribution System | Zheng et al. | 2017 |
| Data Driven Fuzzy Modelling | Rosa et al. | 2017 |
| Data Driven approach for counting apples & oranges | Chen et al. | 2016 |
| Data Driven spoken language understanding system | Yulan et al. | 2003 |
| Health and management | Delesie et al. | 2001 |
| Data quality | Feelders et al. | 2000 |

*Table 1: Data Driven Applications*

| Combination of DD & Knowledge in different applications | Authors | Year |
|---|---|---|
| Combination of knowledge and data driven methods for de-identification of clinical narratives | Dehghan et.al | 2015 |
| Extending knowledge driven activity models through data driven learning techniques | Gorka et al. | 2015 |
| From model, signal to knowledge: A data driven perspective of fault detection and diagnosis | Dai et al. | 2013 |
| A hybrid knowledge based and data driven approach to identifying sementically similar concepts | Pivovarov et al. | 2012 |
| Combining knowledge and data driven insights for identifying risk factors using electronics health records | Sun et al. | 2012 |
| A Dual process model of defence against conscious and unconscious death related thoughts: An extension of terror management theory | Pyszozynski et al. | 1999 |

*Table 2: Combinations of Data Driven and Knowledge*

## Discussion

We discussed different approaches that have been defined in the last few years to overcome the problem of fake news detections in social networks. Most of the approaches based on supervised or unsupervised methods. Those approaches are not providing good results due to non availability of gold standard data set that can help to train and evaluate the classifier and produce good results (Subhabrata et al. 2015). It is the fact that motivations and psychological states of mind of people can be different from the professionals in the real world. Different groups are working now to combat this hot issue and for that purpose they are thinking to utilize actual dataset rather than opinions, blogs. To tackle the problem fake news detection we need to incorporate both behavioral and social entities and to combine knowledge and data. In this chapter we try to discuss all possible types of fake news and impact of that news socially so on the basis of literature evaluation we can say that it is also possible to detect fake news with different known facts like time, location, quality, and stance of others. With these types of measure similarities we can detect the quality of news. In next chapter we will discuss proposed combination statistical analysis of publically available dataset just for understanding the issue more deeply.

## Method

Learning from data and engineered knowledge to overcome fake news issue on social media. To achieve the goal a new combination algorithm approach (Figure-8) shall be developed which will classify the text as soon as the news will publish online.

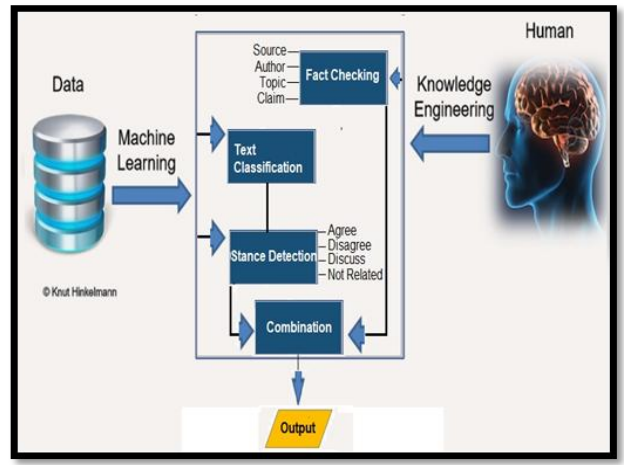

*Figure 8: Block Diagram of the framework*

In developing such a new classification approach as a starting point for the investigation of fake news we first applied publically available data set for our learning. The first step in fake news detection is classifying the text immediately once the news published online. Classification of text is one of the important research issues in the field of text mining. As we knew that dramatic increase in the contents available online gives raise problem to manage this online textual data. So it is important to classify the news into the specific classes (Fake, Non fake, unclear).

Classification of millions of news that published online manually is time consuming and expensive task. So before going to the automatic text classification we need to understand different text classification techniques (Nidhi et al. 2011).

### Selection/Collection of news articles

For training and understanding the classifier we try publically available dataset[13] that is based on the collection of approximate seventeen thousand news articles extracted from online news organizations: Location of the article (country); publication details (Organization, author, date, unique id) Text (Title, full article, online link) & Classification details.

### Data extraction and analysis

The dataset was already sorted qualitatively by the different categories like fake, not fake, bias, conspiracy & hate. Further we classify data with different result indicators (replies, participants, likes, comments and total number of shares). In next step we will show the outcomes of that dataset that will help us to understand the process.

### Results Extracted from dataset and Future goal

The details of the dataset with classified attributes mentioned above in collection tab but here in the figure-9 we just highlighted the results we obtained that how can we specify claims that can be helpful in combination of proposed techniques. From 17946 news articles, 12460 articles were biased category, 572 were fake articles, 870 articles were conspiracy category and 2059 were non-fake articles.

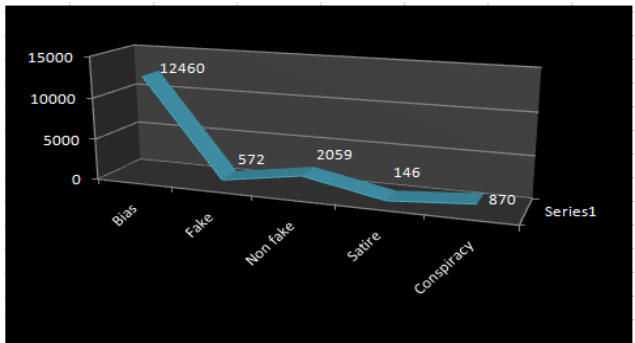

*Figure 9: Comparison results*

Our proposed combination diagram contains two parts data and knowledge part that further classification of these two can be seen in the diagram. Data side contains text classification and stance detection while knowledge side contains fact checking that will help us to refine the results. We categorizer our task in three parts and at the end we will combine the results to check the news status fake or not fake.

## Discussion

---
[13] https://www.kaggle.com/mrisdal/fake-news/version/1

In this section we discussed step by step process that how can we combine learning from data and engineered knowledge in order to combat fake news detection in social networks. Once the news will publish online then classifier will classify the text into the classes fake, non fake and unclear. After text classification then we will check the stance of that particular news which categories the news into four categories agree, disagree, discuss and not related. In the next step we will apply fact checking that will refine our results as fact checking uses engineered knowledge in order to analyze the content of the text and it will compare it to the known facts (see figure).

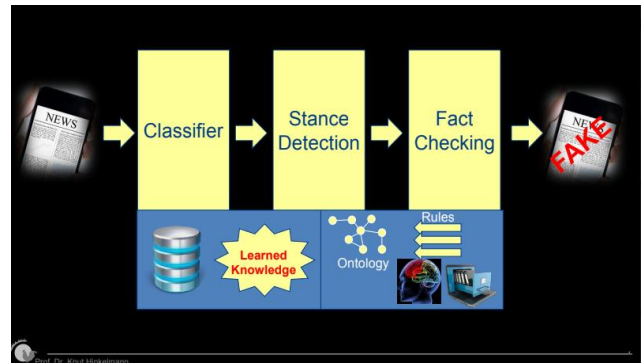

*Figure 10: Proposed Combination of Data Driven and Knowledge*

- When the news published online our proposed classifier will check the similarity between words, text and overall similarity. As per the literature study we have come to know that in news dataset SVM can be good for starting due to its dealing with data as we need to do some mathematical expressions so for that purpose may be we need to use some other library API, s so in those cases it can perform well. Neural network produce good results but if and only if we have large sample size and large storage space. It's also intolerant with noise. Term graph is preferred especially when we have adjacent words and our objective to maintain the correlation between classes. Bayesian classifier can also perform well but only in that case where we have less data set.
- In Stance detection method we will check the viewpoint of the reader of the news is in favor or against or neutral. As per literature there are two ways to represent user stance explicit and implicit. In explicit the readers give direct expressions like thumb up or thumb down. Implicit stances we extracted results from social media.
- Finally we will apply fact checking that will work on two points check worthy claims and discriminate the veracity of claims. We will apply key basis and viewpoints of users on that particular news. Examples of fact checking is open web and structured knowledge graphs.

At the end we will automate our proposed combination that can classify the text automatically and after stance detection and fact checking we will be in the position to get results that the news is fake or not fake.

In this survey, we have covered previous efforts to the development of a fake news system that can detect and resolve the veracity of rumors individually. We discussed in introduction section that fake contents after 2016 Presidential elections it became a big issue and we also knew that the rumor veracity value is unverifiable in the early stages and subsequently resolved as true or false in a relatively short period of time or it can also remain unverified for a long time. We also discussed different detections systems which have distinct characteristics but also commonalities with rumors so it is difficult to detect only with the help of data driven. The approaches discussed in this article are designed to tackle the fake news issue somehow but it is desired that the integration can be helpful for detection (Figure-10).

Since the fake news producers seems to improve their sharing strategies to avoid text classification and detection techniques, fake news detection organizations are required to update their strategies.

## Conclusion

Recently after US presidential elections social media often become a trained vehicle of spread misinformation and hoaxes. No necessary instruments and cognitive abilities required to assess the credibility of the other person just come and share your opinion on social media. May be this one has no serious consequences if only to share or spread rumors of less important but it can be a serious problem when the consumers can purchased products on the basis of these rumors or sometimes serious security issues. Especially in the context of politics that influence public opinion when individuals run small scale or large scale organizations only to ruin someone credibility (e.g., Donald Trump & Hillary Clinton election). In this paper we try to cover the work that includes: knowledge based and style based. Then further we try to explain the sub categories that occur in these two domains e.g. Social context based, propagation based, stance based etc. We try to taken into consideration the effect of fake news on social platforms. We also try to cover some context where false news generates serious issues for the individuals involved. We have presented state of the art block diagram that is the combination of knowledge (Fact-checking) and data (Text classification, Stance detection). As we already discussed that the important open issue is the non availability of a gold standard dataset and predefined benchmark as well as collection of large amounts of fake articles dataset. So on the basis of the points we highlighted one can say that in big data era still the problem has not been received the attention it deserves. But yes few approaches we discussed in expert oriented section that have been proposed automatically asses the fact checking and credibility assessment of news. We can analyze fake news differently with different measure similarities e.g we can detect whether the same news published by other media agencies or not, We can check the location of the news Maybe a news has a higher probability of being fake, if it is generated somewhere else and not at the location they deal with (e.g. Trump writes about China or Arabian States, News about Hillary Clinton has its origin in Russia), We can check news quality wise it is more probable that fake news do not have mentioned their sources, simply claim something, while for real news the source is mention and also we can check the time of the news as whether the same news appears in other media or sourced if it is repeated more often in the beginning, because they are interesting, and become recognized as fake with the time, which reduces the repetition or they are deleted from some websites. At this stage we don't have definitive solution but after detailed literature review we can say that it's true that producing more reports with more facts can be useful for helping us to make such decisions and find technical solutions in fake news detection.

Combination of machine learning and knowledge engineering can be useful for fake news detection as it looks like that fake news may be the most challenging area of research in coming years.